\documentclass[10pt,twocolumn,letterpaper]{article}

\usepackage{cvpr}
\usepackage{times}
\usepackage{epsfig}
\usepackage{graphicx}
\usepackage{amsmath}
\usepackage{amssymb}

\usepackage{multirow}

\newcommand{\w}{\mathbf{w}}
\newcommand{\x}{\mathbf{x}}

\newcommand{\z}{\mathbf{z}}

\newcommand{\I}{\mathbf{I}}

\newcommand{\W}{\mathbf{W}}

\newcommand{\D}{\mathcal{D}}
\newcommand{\G}{\mathcal{G}}

\long\def\comment#1{}

\cvprfinalcopy 

\def\cvprPaperID{1182} 

\ifcvprfinal\fi

\newcommand\Mark[1]{\textsuperscript#1}
\usepackage{authblk}

\begin{document}

\title{Tagging like Humans: Diverse and Distinct Image Annotation}



\author[1]{Baoyuan Wu}
\author[1]{Weidong Chen}
\author[1]{Peng  Sun}
\author[1]{Wei Liu}
\author[2]{Bernard  Ghanem}
\author[3]{Siwei Lyu}

\affil[1]{Tencent AI Lab ~ \Mark{2}KAUST ~ \Mark{3}University at Albany, SUNY 
\authorcr{\tt\small  wubaoyuan1987@gmail.com, powerchen@tencent.com,  pengsun000@gmail.com, wliu@ee.columbia.edu, bernard.ghanem@kaust.edu.sa, slyu@albany.edu}}

\comment{
\author{Baoyuan Wu,  Weidong Chen, Peng  Sun, Wei Liu, Bernard  Ghanem, Siwei Lyu}
\affil{Tencent AI Lab,  Tencent AI Lab, Tencent AI Lab, Tencent AI Lab, KAUST, University at Albany, SUNY \authorcr{wubaoyuan1987@gmail.com}
}
}

%

\maketitle
\thispagestyle{empty}

\begin{abstract}
In this work we propose a new automatic image annotation model, dubbed {\bf diverse and distinct image annotation} (D$^2$IA). The generative model D$^2$IA is inspired by the ensemble of human annotations, which create semantically relevant, yet distinct and diverse tags. 
In D$^2$IA, we generate a relevant and distinct tag subset, in which the tags are relevant to the image contents and semantically distinct to each other, using sequential sampling from a determinantal point process (DPP) model.
Multiple such tag subsets that cover diverse semantic aspects or diverse semantic levels of the image contents are generated by randomly perturbing the DPP sampling process.
We leverage a generative adversarial network (GAN) model to train D$^2$IA. 
Extensive experiments including quantitative and qualitative comparisons, as well as human subject studies, on two benchmark datasets demonstrate that the proposed model can produce more diverse and distinct tags than the state-of-the-arts. 
\end{abstract}

\section{Introduction}
\label{sec: introduction}

\begin{figure*}[t]
\centering
\includegraphics[width=0.97\textwidth,height=3.3in]{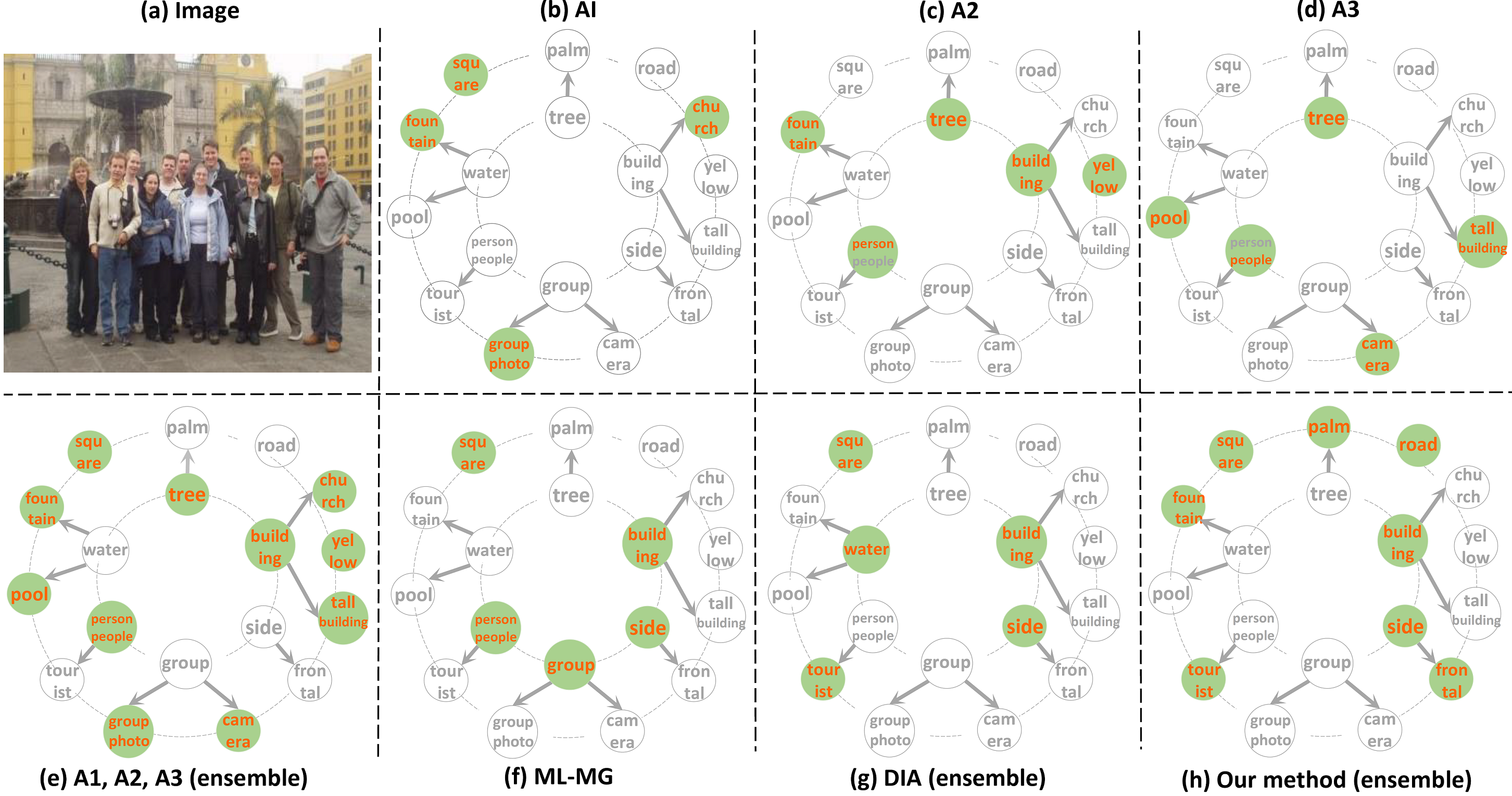}
\caption{{\bf An example illustrating the diversity and distinctiveness in image annotation}. The image \textbf{(a)} is from IAPRTC-12 \cite{iaprtc-12-data-2006}.
We present the tagging results from 3 independent human annotators \textbf{(v)}-\textbf{(d)}, identified as A1, A2, A3, respectively, as well as their ensemble result \textbf{(e)}. 
We also present the results of some automatic annotation methods. ML-MG \cite{my-iccv-2015} \textbf{(f)} is a standard annotation method that requires the relevant tags. 
DIA (ensemble) \cite{my-cvpr-2017-dia} \textbf{(g)} indicates that we repeat the sampling of DIA for 3 times, with the requirement that each subset includes at most 5 tags, and then combine these 3 subsets to one ensemble subset. 
Similarly, we obtain the ensemble subset of our method \textbf{(h)}.   
In each graph, nodes are candidate tags and the arrows connect parent and child tags in the semantic hierarchy. 
This figure is better viewed in color.}
\label{fig-1}
\vspace{-1em}
\end{figure*}

Image annotation is one of the fundamental tasks of computer vision with many applications in image retrieval, caption generation and visual recognition. Given an input image, an image annotator outputs a set of keywords (tags) that are relevant to the content of the image.  Albeit an impressive progress has been made by current image annotation algorithms, to date, most of them \cite{LEML-ICML-2014, my-iccv-2015, pairwise-ranking-jiebo-cvpr-2017} focus on the {\em relevancy} of the obtained tags to the image with little consideration to their inter-dependencies. As a result, algorithmically generated tags for an image are relevant but at the same time less informative, with redundancy among the obtained tags, {\it e.g.}, one state-of-the-art image annotation algorithm ML-MG \cite{my-iccv-2015} generates tautology {\it`people'} and {\it `person'} for the image in Fig. \ref{fig-1}(f). 

This is different from how human annotators work. We illustrate this using an annotation task involving three human annotators (identified as A1,A2 and A3). Each annotator was asked to independently annotate the first $1,000$ test images in the IAPRTC-12 dataset \cite{iaprtc-12-data-2006} with the requirement of ``describing the main contents of one image using as few tags as possible".  
One example of the annotation results is presented in Fig. \ref{fig-1}. Note that individual human annotators tend to use semantically {\em distinct} tags (see Fig.  \ref{fig-1} (b)-(d)), and the semantic redundancy among tags is lower than that among the tags generated by the annotation algorithm ML-MG \cite{my-iccv-2015} (see Fig. \ref{fig-1}(f)).  
Improving the semantic distinctiveness of generated tags has been studied in recent work \cite{my-cvpr-2017-dia}, which uses a determinant point process (DPP) model \cite{dpp-for-machine-learning-2012} to produce tags with less semantic redundancies.
The annotation result of running this algorithm on the example image is shown in Fig. \ref{fig-1}(g).

However, such results still lack in one aspect when comparing with the annotations from the ensemble of human annotators (see Fig. \ref{fig-1}(e)). The collective annotations from human annotators also tend to be {\em diverse}, consisting of tags that cover more semantic elements of the image. For instance, different human annotators tend to use tags across different abstract levels, such as {\it `church'} vs. {\it `building'}, to describe the image. Furthermore, different human annotators usually focus on different parts or elements of the image. For example, A1 describes the scene as {\it `square'}, A2 notices the {\it `yellow'} color of the building, while A3 finds the {\it `camera'} worn on the chest of people.

\begin{figure}[t]
\centering
\includegraphics[width=0.47\textwidth,height=1.5in]{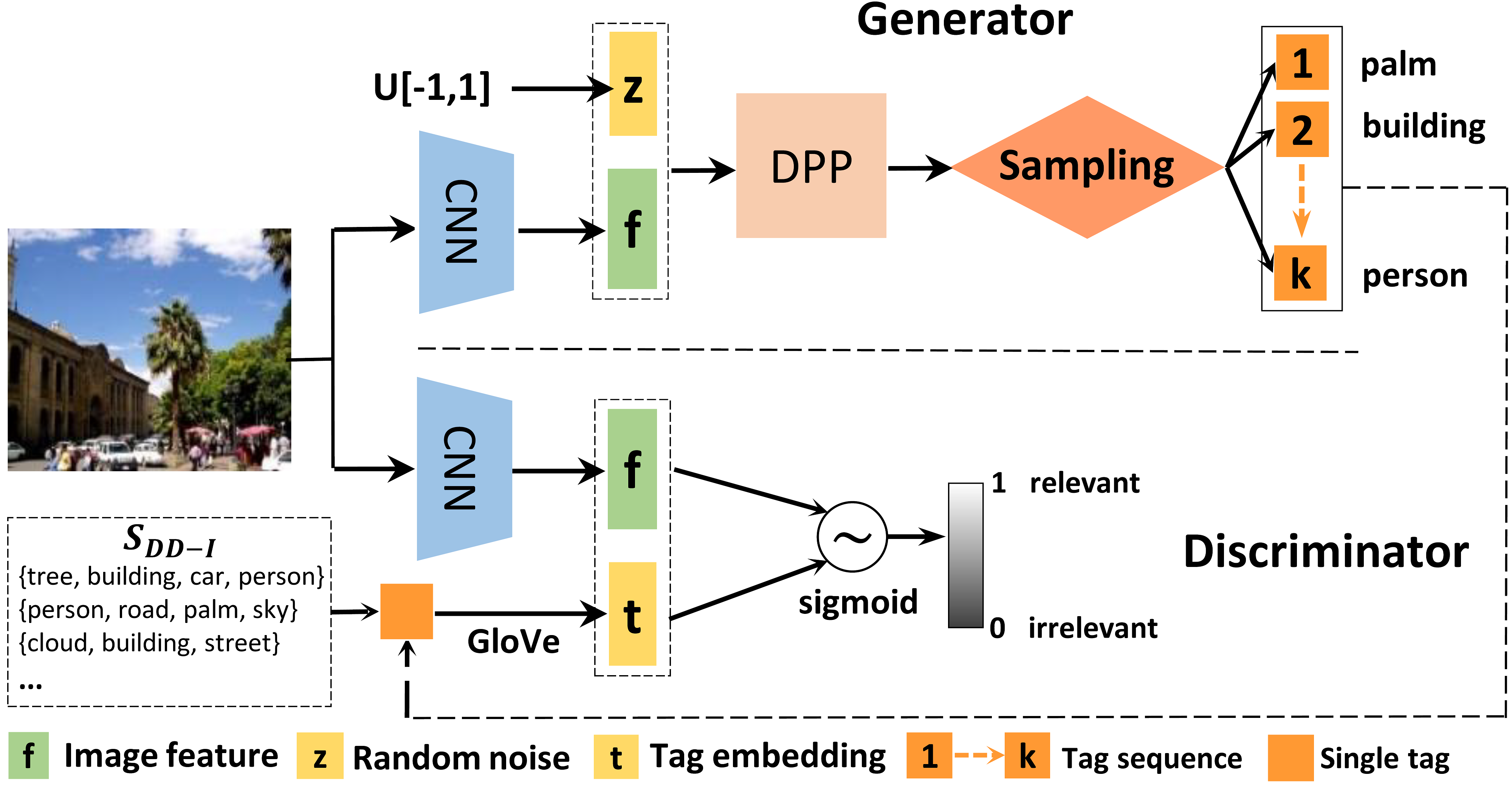}
\caption{A schematic illustration of the structure of the proposed D$^2$IA-GAN model. $S_{DD-I}$ indicates the ground-truth set of diverse and distinct tag subsets for the image $I$, which will be defined in the Section \ref{sec: background}.}
\label{fig-2}
\vspace{-.2in}
\end{figure}

In this work, we propose a novel image annotation model, namely diverse and distinct image annotation (D$^2$IA), which aims to improve the diversity and distinctiveness of the tags for an image by learning a generative model of tags from multiple human annotators. 
The distinctiveness enforces the semantic redundancy among the tags in the same subset to be small, while the diversity encourages different tag subsets to cover different aspects or different semantic levels of the image contents. 
Specifically, this generative model first maps the concatenation of the image feature vector and a random noise vector to a posterior probability with respect to all candidate tags, and then incorporates it into a determinantal point process (DPP) model \cite{dpp-for-machine-learning-2012} to generate a distinct tag subset by sequential sampling. 
Utilizing multiple random noise vectors for the same image, multiple diverse tag subsets are sampled. 

We train D$^2$IA as the generator in a generative adversarial network (GAN) model \cite{gan-nips-2014} given a large amount of human annotation data, which is subsequently referred to as 
D$^2$IA-GAN. 
The discriminator of D$^2$IA-GAN is a neural network measuring the relevance between the image feature and the tag subset that aims to distinguish the generated tag subsets and the ground-truth tag subsets from human annotators. 
The general structure of D$^2$IA-GAN model is shown in Fig. \ref{fig-2}.
The proposed D$^2$IA-GAN is trained by alternative optimization of the generator and discriminator while fixing the other until convergence.

One characteristic of the D$^2$IA-GAN model is that its generator includes a sampling step which is not easy to optimize directly using gradient based optimization methods. Inspired by reinforcement learning algorithms, we develop a method based on the {\it policy gradient} (PG) algorithm, where we model the discrete sampling with a differentiable policy function (a neural network), and devise a {\it reward} to encourage the generated tag subset to match the image content as close as possible. 
Incorporating the policy gradient algorithm in the training of D$^2$IA-GAN, we can effectively obtain the generative model for tags conditioned on the image. 
As shown in Fig. \ref{fig-1}(h), using the trained generator of D$^2$IA-GAN can produce 
diverse and distinct tags that are closer to those generated from the ensemble of multiple human annotators (Fig. \ref{fig-1}(e)).

The main contributions of this work are four-fold. 
{\bf (1)} We develop a new image annotation method, namely diverse and distinct image annotator (D$^2$IA), to create relevant, yet distinct and diverse annotations for an image, which are more similar to tags provided by different human annotators for the same image;
{\bf (2)} we formulate the problem as learning a probabilistic generative model of tags conditioned on the image content, which exploits a DPP model to ensure distinctiveness and conducts random perturbations to improve diversity of the generated tags; 
{\bf (3)} the generative model is adversarially trained using a specially designed GAN model that we term as D$^2$IA-GAN;
{\bf (4)} in the training of D$^2$IA-GAN we use the policy gradient algorithm to handle the discrete sampling process in the generative model. 
We perform experimental evaluations on ESP Game \cite{espgame-2004} and IAPRTC-12 \cite{iaprtc-12-data-2006} image annotation datasets, and subject studies based on human annotators for the quality of the generated tags. The evaluation results show that the tag set produced by D$^2$IA-GAN is more diverse and distinct when comparing with those generated by the state-of-the-art methods.

\section{Related Work}
\label{sec: related work}

Existing image annotation methods fall into two general categories: they either generate all tags simultaneously using multi-label learning, or predict tags sequentially using sequence generation.
The majority of existing image annotation methods are in the first category.
They mainly differ in designing different loss functions or exploring different class dependencies. 
Typical loss functions include square loss \cite{LEML-ICML-2014, muitilabel-attention-cvpr-2017},  
ranking loss \cite{cnn-image-annotation-arxiv-2013, pairwise-ranking-jiebo-cvpr-2017}, cross-entropy loss \cite{spatial-regularization-cvpr-2017}), {\it etc.} 
Commonly used class dependencies include class co-occurrence \cite{my-icpr-2014,my-pr-2015,li2016facial}, mutual exclusion \cite{xiaotong-multi-label-exclusive-2011, deep-dpp-cvpr-2017}, class cardinality \cite{my-aaai-2016-imbalance}, sparse and low rank \cite{my-ijcv-2018}, and semantic hierarchy \cite{my-iccv-2015}. 
Besides, some multi-label learning methods consider different learning settings, such as multi-label learning with missing labels \cite{my-icpr-2014, my-pr-2015}, label propagation in semi-supervised learning \cite{ssl-wei-liu-icml-2010, teach-to-learn-wei-liu-aaai-2016, teach-to-learn-wei-liu-tnnls-2017} and transfer learning \cite{multi-label-transfer-wei-liu-pami-2017} settings. 
A thorough review of multi-label learning based image annotation methods can be found in      
 \cite{review-image-annotation-pr-2012}.

Our method falls into the second category, which generates tags in a sequential manner. This can better employ the inter-dependencies of the tags. 
Many methods in this category are built on sequential models, such as recurrent neural networks (RNNs), which work in coordination with convolutional neural networks (CNNs) to exploit their representation power for images. The main difference of these works lies in designing an interface between CNN and RNN.
In \cite{rnn-image-annotation-icpr-2016}, features extracted by a CNN model were used as the hidden states of a RNN. 
In \cite{rnn-cnn-image-annotation-cvpr-2016}, the CNN features were integrated with the output of a RNN. 
In \cite{rnn-semantic-regularization-cvpr-2017}, the predictions of a CNN were used as the hidden states of a RNN, and the ground-truth tags of images were used to supervise the training of the CNN. 
Not directly using the output layer of a RNN, the work in \cite{rnn-fisher-vector-eccv-2016} utilized the Fisher vector derived from the gradient of the RNN, as the feature representation.

Although RNN is a suitable model for the sequential image annotation
task for its ability to implicitly encode the dependencies
among tags, it is not easy to explicitly embed
some prior knowledge about the tag dependencies like semantic
hierarchy \cite{my-iccv-2015} or mutual exclusion \cite{xiaotong-multi-label-exclusive-2011} in the RNN model. 
To remedy this issue, the recent work of DIA \cite{my-cvpr-2017-dia} formulated the sequential prediction as a sampling process based on a determinantal point process (DPP) \cite{dpp-for-machine-learning-2012}. 
DIA encodes the class co-occurrence into the learning process, and incorporates the semantic hierarchy into the sampling process. 
Another important difference between
DIA and the RNN-based methods is that the former explicitly
embeds the negative correlations among tags {\it i.e.}, avoiding using semantically similar tags for the same image, while RNN-based methods typically ignore such negative corrlations.
The main reason is that the objective of DIA is
to describe an image with a few diverse and relevant tags,
while most other methods tend to predict most relevant tags.

Our proposed model D$^2$IA-GAN is inspired by DIA, and both are developed based on the observations of human annotations. Yet, there are several significant differences between them. 
The most important difference is in their objectives. 
DIA aims to simulate a single human annotator to use semantically distinct
tags for an image, while D$^2$IA-GAN aims to simulate multiple human annotators simultaneously to capture the diversity among human annotators. 
They are also different in the training process, which will be reviewed in the Section \ref{sec: model}. 
Besides, in DIA \cite{my-cvpr-2017-dia}, `diverse/diversity' refers to the semantic difference between tags in the same tag subset, to which we use the word `distinct/distinctiveness' for the same meaning in this work. We use `diverse/diversity' to indicate the semantic difference between multiple tag subsets for the same image.

\section{Background}
\label{sec: background}

\begin{figure}[t]
\centering
\includegraphics[width=0.43\textwidth,height=1.0in]{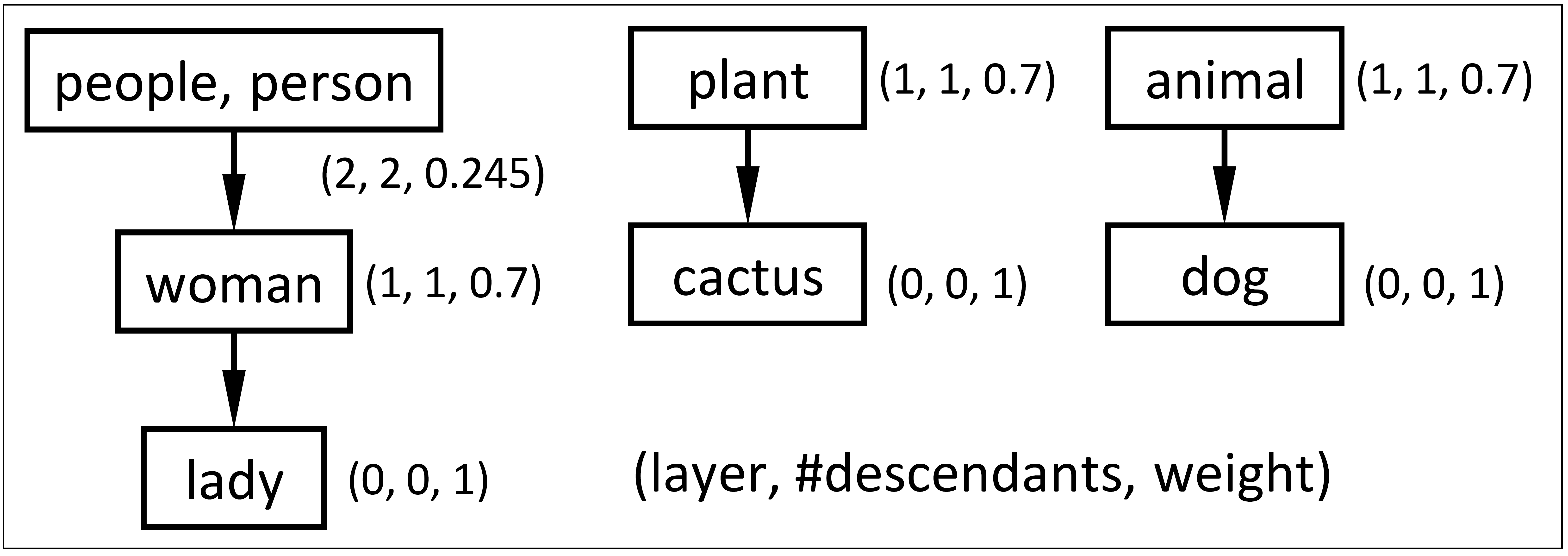}
\caption{A brief example of the weighted semantic paths. The word in box indicates the tag. The arrow $a \rightarrow b$ tells that tag $a$ is the semantic parent of tag $b$. The bracket close to each box denotes the corresponding (node layer, number of descendants, tag weight). Boxes connected by arrows construct a semantic path.}
\label{fig-3-weighted-path}
\vspace{-.2in}
\end{figure}

\noindent
{\bf Weighted semantic paths.} Weighted semantic paths \cite{my-cvpr-2017-dia} are constructed based on the semantic hierarchy and synonyms \cite{my-iccv-2015} among all candidate tags. To construct a weighted semantic path, we treat each tag as a node, and the synonyms are merged into one node. Then, starting from each leaf node in the semantic hierarchy, we connect its direct parent node and repeat this connection process, until the root node is achieved. All tags that are visited in this process form the weighted semantic path of the leaf tag. The weight of each tag in the semantic path is computed inversely proportional to the node layer (the layer number starts from 0 at leaf nodes) and the number of descendants of each node. 
As such, the weight of the tag with more specified information will be larger.
A brief example of the weighted semantic paths is shown in Fig. \ref{fig-3-weighted-path}. 
We use $SP_{\mathcal{T}}$ to denote the semantic paths of set $\mathcal{T}$ of all candidate tags.  
$SP_T$ indicates the semantic paths of the tag subset $T$. 
$SP_I$ represents the weighted semantic paths of all ground-truth tags of image $I$. 

\vspace{3pt}
\noindent
{\bf Diverse and distinct tag subsets.} Given an image $I$ and its ground-truth semantic paths $SP_I$, a tag subset is distinct if there are no tags being sampled from the same semantic path. 
An example of the distinct tag subset is shown in Fig. \ref{fig-3-weighted-path}:  $SP_I = \{ lady \rightarrow woman \rightarrow (people, person), cactus \rightarrow plant \}$ includes 3 semantic paths with 7 tags, such as $\{lady, plant\}$ or $\{women, cactus\}$. 
A tag set is diverse if it includes multiple distinct tag subsets. These subsets cover different contents of the image, due to two possible reasons, including 1) they describe  different contents of the image, and 2) they describe the same content but at different semantic levels.  
As shown in Fig. \ref{fig-3-weighted-path}, we can construct a diverse set of distinct tag subsets like $\{ \{lady, cactus \}, \{ plant, cat \}, \{woman, plant, animal\}  \}$. 
Furthermore, we can construct all possible distinct tag subsets (ignoring the subset with a single  tag) to obtain the complete diverse set of distinct tag subsets, referred to as $S_{DD-I}$. 
Specifically, for the subset with 2 tags, we will pick 2 paths out of 3 and sample one tag from each picked path. Then we obtain in total 16 distinct subsets. For the subset with 3 tags, we sample one tag from each semantic path, leading to 12 distinct subsets. 
$S_{DD-I}$ will be used as the ground-truth to train the proposed model.  

\vspace{3pt}
\noindent
{\bf Conditional DPP.} 
We use a conditional determinantal point process (DPP) model to measure the probability of the tag subset $T$, derived from the ground set $\mathcal{T}$ given a feature $\x$ of the image $I$. The DPP model is formulated as 
\begin{flalign}
\vspace{-0.1in}
 \mathcal{P}(T|I) = \frac{\text{det}\big(\mathbf{L}_{T}(I)\big)}{\text{det}\big(\mathbf{L}_{\mathcal{T}}(I) + \I\big)},
 \label{eq: formulation of conditional DPP}
 \vspace{-0.13in}
\end{flalign}
where $\mathbf{L}_{\mathcal{T}}(I) \in \mathbb{R}^{|\mathcal{T}| \times |\mathcal{T}|}$ is a positive semi-definite kernel matrix. 
$\I$ indicates the identity matrix. 
For clarity, the parameters of $\mathbf{L}_{\mathcal{T}}(I)$ and (\ref{eq: formulation of conditional DPP}) have been omitted. 
The sub-matrix $\mathbf{L}_{T}(I) \in \mathbb{R}^{|T| \times |T|}$ is constructed by extracting the rows and columns corresponding to the tag indexes in $T$. 
For example, assuming $\mathbf{L}_{\mathcal{T}}(I) = [ a_{ij} ]_{i,j = 1, 2, 3, 4}$ and $T = \{2,4\}$, then $\mathbf{L}_{T}(I) = [a_{22}, a_{24}; a_{42}, a_{44}]$. 
$\text{det}\big(\mathbf{L}_{T}(I)\big)$ indicates the determinant of $\mathbf{L}_{\mathcal{T}}(I)$. It encodes the negative correlations among the tags in the subset $T$. 

Learning the kernel matrix $\mathbf{L}_{\mathcal{T}}(I)$ directly is often difficult, especially when $|\mathcal{T}|$ is large. To alleviate this problem, we decompose $\mathbf{L}_{\mathcal{T}}(I)$ as $\mathbf{L}_{\mathcal{T}}(i,j) = v_i \boldsymbol{\phi}_i^\top \boldsymbol{\phi}_i v_j$, 
where the scalar $v_i$ indicates the individual score with respect to tag $i$, and $\mathbf{v}_{\mathcal{T}} = [ v_1, \ldots, v_i, \ldots, v_{\mathcal{T}} ]$. 
The vector $\boldsymbol{\phi}_i \in \mathbb{R}^{d'}$ corresponds to the direction of tag $i$, with $\parallel\boldsymbol{\phi}_i\parallel = 1$, and can be used to construct the semantic similarity matrix $\mathbf{S}_{\mathcal{T}} \in \mathbb{R}^{|\mathcal{T}| \times |\mathcal{T}|}$ with $\mathbf{S}_{\mathcal{T}}(i,j) = \boldsymbol{\phi}_i^\top \boldsymbol{\phi}_j$.  
With this decomposition, we can learn $\mathbf{v}_{\mathcal{T}}$ and  $\mathbf{S}_{\mathcal{T}}$ separately. 
More details of DPP can be found in \cite{dpp-for-machine-learning-2012}. 
In this work, $\mathbf{S}_{\mathcal{T}}$ is pre-computed as:
\begin{flalign}
\vspace{-0.13in}
\mathbf{S}_{\mathcal{T}}(i,j) = \frac{1}{2} + \frac{\langle \mathbf{t}_i, \mathbf{t}_j\rangle }{ 2\| \mathbf{t}_i \|_2 \|\mathbf{t}_j\|_2} \in [0, 1]~~\forall~i,j\in\mathcal{T},
\vspace{-0.13in}
\end{flalign}
where the tag representation $\mathbf{t}_i \in \mathbb{R}^{50}$ is derived from the GloVe algorithm \cite{glove-2014}. $\langle \cdot, \cdot \rangle$ indicates the inner product of two vectors, while $\|\cdot\|_2$ denotes the $\ell_2$ norm of a vector.

\vspace{3pt}
\noindent
{\bf k-DPP sampling with weighted semantic paths.}
k-DPP sampling \cite{dpp-for-machine-learning-2012} is a sequential sampling process to obtain a tag subset $T$ with at most $k$ tags, according to the distribution (\ref{eq: formulation of conditional DPP}) and the weighted semantic paths $SP_{\mathcal{T}}$. It is denoted as $\mathcal{S}_{\text{k-DPP}, SP_{\mathcal{T}}}(\mathbf{v}_{\mathcal{T}}, \mathbf{S}_{\mathcal{T}})$ subsequently. 
Specifically, in each sampling step, the newly sampled tag will be checked whether it is from the same semantic path with any previously sampled tags. If not, it is included into the tag subset; if yes, it is abandoned and we go on sampling the next tag, until $k$ tags are obtained. 
The whole sampling process is repeated multiple times to obtain different tag subsets. 
Then the subset with the largest tag weight summation is picked as the final output. 
Note that a larger weight summation indicates more semantic information. Since the tag weight is pre-defined when introducing the weighted semantic paths, it is an objective criterion to pick the subset.

\section{D$^2$IA-GAN Model}
\label{sec: model}

Given an image $I$, we aim to generate a diverse tag set including multiple distinct tag subsets relevant to the image content, as well as an ensemble tag subset of these distinct subsets, which could provide a comprehensive description of $I$.
These tags are sampled from a generative model conditioned on the image, and we use a  conditional GAN (CGAN) \cite{cgan-2014, wenhan-cvpr-2018, face-gan} to train it, with the generator part $\G$ being our model and a discriminator $\D$, as shown in Fig. \ref{fig-2}.
Specifically, conditioned on $I$, $\G$ projects one noise vector $\z$ to one distinct tag subset $T$, and uses different noise vectors to ensure diverse/different tag subsets. 
$\D$ serves as an adversary of $\G$, aiming to distinguish the generated tag subsets using $\G$ from the ground-truth ones $S_{DD-I}$.  

\subsection{Generator}

The tag subset $T \subset \mathcal{T} = \{1, 2, \ldots, m\}$ with $|T| \leq k$ can be generated from the generator $G_{\boldsymbol{\theta}}(I, \mathbf{z})$, according to the input image $I$ and a noise vector $\z$, as follows:
\begin{flalign}
\hspace{-0.7em} \G_{\boldsymbol{\theta}}(I, \mathbf{z}; \mathbf{S}_{\mathcal{T}}, SP_{\mathcal{T}}, k) 
\sim
 \mathcal{S}_{\text{k-DPP}, SP_{\mathcal{T}}}\big( \sqrt{\mathbf{q}_{\mathcal{T}}(I, \z)}, \mathbf{S}_{\mathcal{T}} \big).
\label{eq: G model} 
\end{flalign}
%
The above generator is a composite function with two parts. 
The {\bf inner} part $\mathbf{q}_{\mathcal{T}}(I, \z) = \sigma\big(\W^\top [f_{\G}(I); \z]+\mathbf{b}_{\G} \big) \in [0,1]^{|\mathcal{T}|}$ is a CNN based soft classifier. 
$f_{\G}(I)$ represents the output vector of the fully-connected layer of a CNN model, and $[\mathbf{a}_1; \mathbf{a}_2]$ denotes the concatenation of two vectors $\mathbf{a}_1$ and $\mathbf{a}_2$. $\sigma(\mathbf{a}) = \frac{1}{1+\exp(-\mathbf{a})}$ is the sigmoid function. $\sqrt{\mathbf{a}}$ indicates the element-wise square root of vector $\mathbf{a}$.
The parameter matrix $\W = [\w_1, \ldots, \w_i, \ldots, \w_m] \in \mathbb{R}^{m \times d}$ and the bias parameter $\mathbf{b}_{\G} \in \mathbb{R}^{d}$ map the feature vector $[f_{\G}(I); \z] \in \mathbb{R}^{d}$ to the logit vector.  
The trainable parameter $\boldsymbol{\theta}$ includes $\W, \mathbf{b}_{\G}$ and the parameters of $f_{\G}$.
The noise vector $\z$ is sampled from the uniform distribution $\text{U}[-1,1]$. 
The {\bf outer} part $\mathcal{S}_{\text{k-DPP}, SP_{\mathcal{T}}}(\sqrt{\mathbf{q}_{\mathcal{T}}(I, \z)}, \mathbf{S}_{\mathcal{T}})$ is the k-DPP sampling with weighted semantic paths $SP_{\mathcal{T}}$ (see Section \ref{sec: background}). Using $\sqrt{\mathbf{q}_{\mathcal{T}}(I, \z)}$ as the quality term and utilizing the pre-defined similarity matrix $\mathbf{S}_{\mathcal{T}}$, then a conditional DPP model can be constructed as described in Section \ref{sec: background}.

\subsection{Discriminator}

$\D_{\boldsymbol{\eta}}(I, T)$ evaluates the relevance of image $I$ and tag subset $T$: it outputs a value in $[0,1]$, with $1$ meaning the highest relevance and $0$ being the least relevant. 
Specifically, $\D_{\boldsymbol{\eta}}$ is constructed as follows: first, as described in Section \ref{sec: background}, each tag $i \in T$ is represented by a vector $\boldsymbol{t}_i \in \mathbb{R}^{50}$ derived from the GloVe algorithm \cite{glove-2014}. Then, we formulate $\D_{\boldsymbol{\eta}}(I, T)$ as
\vspace{-0.7em}
\begin{eqnarray}
\vspace{-8em}
\D_{\boldsymbol{\eta}}(I, T) = \frac{1}{|T|}\sum_{i \in T} \sigma\left( \w_{\D}^\top [ f_{\D}(I) ; \boldsymbol{t}_i ] + b_{\D} \right),
\vspace{-8em}
\label{eq: D model}
\vspace{-8em}
\end{eqnarray}
where $f_{\D}(I)$ denotes the output vector of the fully-connected layer of a CNN model  (different from that used in the generator). 
$\boldsymbol{\eta}$ includes $\w_{\D} \in \mathbb{R}^{|f_{\D}(I)|+50}, b_{\D} \in \mathbb{R}$ and the parameters of $f_{\D}(I)$ in the CNN model.

\subsection{Conditional GAN}
\label{sec: subsec conditional GAN}

Following the general training procedure, we learn D$^2$IA-GAN by iterating two steps until convergence: 
(1) fixing the discriminator $\D_{\boldsymbol{\eta}}$ and optimizing the generator $\G_{\boldsymbol{\theta}}$ using (\ref{eq: sub-problem G}), as shown in Section \ref{sec: subsec optimizing G}; 
(2) fixing $\G_{\boldsymbol{\theta}}$ and optimizing $\D_{\boldsymbol{\eta}}$ using (\ref{eq: maximizing discriminator with F1}), as shown in Section \ref{sec: subsec optimizing D}.

\vspace{-0.6em}
\subsubsection{Optimizing $\G_{\boldsymbol{\theta}}$} 
\label{sec: subsec optimizing G}

Given $\D_{\boldsymbol{\eta}}$, we learn $\G_{\boldsymbol{\theta}}$ by 
\vspace{-0.6em}
\begin{eqnarray}
\vspace{-0.8em}
\underset{\boldsymbol{\theta}}{\min} ~
\mathbb{E}_{\mathbf{z} \sim \text{U}[-1,1]} \left[\log \bigg( 1 -  D_{\boldsymbol{\eta}}\big(I, \G_{\boldsymbol{\theta}}(I, \mathbf{z})\big) \bigg) \right].
\vspace{-0.8em}
\label{eq: sub-problem G}
\vspace{-0.8em}
\end{eqnarray}
For clarity, we only show the case with one training image $I$ in the above formulation.
Due to the discrete sampling process $\mathcal{S}(\mathbf{v}_{\mathcal{T}}, \mathbf{S}_{\mathcal{T}})$ in $\G_{\boldsymbol{\theta}}(I, \mathbf{z})$, we cannot optimize (\ref{eq: sub-problem G}) using any existing continuous optimization algorithm. 
To address this issue, we view the sequential generation of tags as controlled by a continuous policy function, which weighs different choices of the next tag based on the image and tags already generated. As such, we can use the policy gradient (PG) algorithm in reinforcement learning for its optimization.
Given a sampled tag subset $T_{\G}$ from $\mathcal{S}(\mathbf{v}_{\mathcal{T}}, \mathbf{S}_{\mathcal{T}})$, the original objective function of (\ref{eq: sub-problem G}) is approximated by a continuous function. 
Specifically, we denote $T_{\G} = \{y_{[1]}, y_{[2]}, \ldots, y_{[k]}\}$, where $[i]$ indicates the sampling order, and its subset $T_{\G-i} = \{y_{[1]}, \ldots, y_{[i]}\}, i \leq k$ includes the first $i$ tags in $T_{\G}$. 
Then, with an instantialized $\z$ sampled from $[-1,1]$, the approximated function is formulated as
\vspace{-0.8em}
\begin{flalign}
\hspace{-0.8em}\mathcal{J}_{\boldsymbol{\theta}}(T_{\G}) & =  
\sum_{i=1}^k \mathcal{R}(I, T_{\G-i}) \log\bigg( \hspace{-0.3em} \prod_{t_1 \in T_{\G-i}} \hspace{-0.75em} q_{t_1}^1 \hspace{-0.25em} \prod_{t_2 \in \mathcal{T} \setminus T_{\G-i}} \hspace{-0.8em} q_{t_2}^0  \bigg), 
\vspace{-0.8em}
\label{eq: objective of policy gradient}
\vspace{-0.8em}
\end{flalign}
where $\mathcal{T} \setminus T_{\G-i}$ denotes the relative complement of $T_{\G-i}$ with respect to $\mathcal{T}$. 
$q_{t}^1 = \sigma\big(\w_t^\top [f_{\G}(I); \z] + b_{\G}(t)\big)$ indicates the posterior probability, and $q_{t}^0 = 1- q_{t}^1$. 
The reward function $\mathcal{R}(I, T_{\G})$ encourages the content of $I$ and the tags $T_{\G}$ to be consistent, and is defined as
\vspace{-0.8em}
\begin{flalign}
\mathcal{R}(I, T_{\G}) & = -\log\big( 1 -  \D_{\boldsymbol{\eta}}(I, T_{\G}) \big).
\vspace{-0.8em}
\label{eq: reward}
\vspace{-0.8em}
\end{flalign}
Compared to a full PG objective function, in (\ref{eq: objective of policy gradient}) we have replaced the \emph{return} with the \emph{immediate reward} $\mathcal{R}(I, T_{\G})$, and the \emph{policy} probability with the decomposed likelihood $\prod_{t_1 \in T_{\G-i}}  q_{t_1}^1 \prod_{t_2 \in \mathcal{T} \setminus T_{\G-i}} q_{t_2}^0$.
Consequently, it is easy to compute the gradient $\frac{\partial \mathcal{J}_{\boldsymbol{\theta}}(T_{\G})}{\partial \boldsymbol{\theta}}$, which will be used in the stochastic gradient ascent algorithm and back-propagation \cite{back-propagation-hinton-1986} to update $\boldsymbol{\theta}$. 

When generating $T_{\G}$ during training, we repeat the sampling process multiple times to obtain different subsets. Then, as the ground-truth set $S_{DD-I}$ for each training image is available, the semantic F$_{1-sp}$ score (see Section \ref{sec: experiments}) for each generated subset can  be computed, and the one with the largest F$_{1-sp}$ score will be used to update parameters. This process encourages the model to generate tag subsets more consistent with the evaluation metric.

\subsubsection{Optimizing $\D_{\boldsymbol{\eta}}$}
\label{sec: subsec optimizing D}

Utilizing the generated tag subset $T_{\G}$ from the fixed generator 
$\G_{\boldsymbol{\theta}}(I, \mathbf{z})$, we learn $\D_{\boldsymbol{\eta}}$ by  
\vspace{-0.6em}
\begin{flalign}
& \underset{\boldsymbol{\eta}}{\max}  ~
\frac{1}{|S_{DD-I}|}\sum_{T \in S_{DD-I} } \bigg[ \beta \log \D_{\boldsymbol{\eta}}(I, T) - (1-\beta) \cdot
\label{eq: maximizing discriminator with F1}
\\
&  \big(\D_{\boldsymbol{\eta}}(I, T) - F_{1-sp}(I, T) \big)^2 \bigg] + \beta \log\big( 1 -  \D_{\boldsymbol{\eta}}(I, T_{\G}) \big) -
\nonumber
\\
&  (1-\beta) \left(\D_{\boldsymbol{\eta}}(I, T_{\G}) - \text{F}_{1-sp}(I, T_{\G})\right)^2,
\nonumber
\vspace{-0.8em}
\end{flalign}
where semantic score $\text{F}_{1-sp}(I, T)$ measures the relevance between the tag subset $T$ and the content of $I$. 
If we set the trade-off parameter $\beta=1$, then (\ref{eq: maximizing discriminator with F1}) is equivalent to the objective used in the standard GAN model. 
For $\beta \in (0,1)$, (\ref{eq: maximizing discriminator with F1}) also encourages the updated $\D_{\boldsymbol{\eta}}$ to be close to the semantic score   $F_{1-sp}(I, T)$. 
We can then compute the gradient of (\ref{eq: maximizing discriminator with F1}) with respect to $\boldsymbol{\eta}$, and use the stochastic gradient ascent algorithm and back-propagation \cite{back-propagation-hinton-1986} to update $\boldsymbol{\eta}$.

\section{Experiments}
\label{sec: experiments}

\subsection{Experimental Settings}

\noindent
{\bf Datasets.} 
We adopt two benchmark datasets, ESP Game \cite{espgame-2004} and IAPRTC-12 \cite{iaprtc-12-data-2006} for evaluation. One important reason for choosing these two datasets is that they have complete weighted semantic paths of all candidate tags $SP_{\mathcal{T}}$, the ground-truth weighted semantic paths of each image $SP_{I}$, the image features and the trained DIA model, provided by the authors of \cite{my-cvpr-2017-dia} and available on GitHub\footnote{Downloaded from {\it https://github.com/wubaoyuan/DIA}}. 
%
Since the weighted semantic paths are important to our method, these two datasets facilitate its evaluation.
Specifically, in ESP Game, there are 18689 train images, 2081 test images, 268 candidate classes, 106 semantic paths corresponding to all candidate tags, and the feature dimension is 597; in IAPRTC-12, there are 17495 train images, 1957 test images, 291 candidate classes, 139 semantic paths of all candidate tags, and the feature dimension is 536.

\vspace{3pt}
\noindent
{\bf Model training.} 
We firstly fix the CNN models in both $\G_{\boldsymbol{\theta}}$ and $\D_{\boldsymbol{\eta}}$ as the VGG-F model\footnote{Downloaded from {\it http://www.vlfeat.org/matconvnet/pretrained/}} pre-trained on ImageNet \cite{imagenet-cvpr-2009}. Then we initialize the columns of the fully-connected parameter matrix $\W$ (see Eq. (\ref{eq: G model})) that corresponds to the image feature $f_{\G}(I)$ using the trained DIA model, while the columns corresponding to the noise vector $\z$ and the bias parameter $\mathbf{b}_{\G}$ are randomly initialized. 
We pre-train $\D_{\boldsymbol{\eta}}$ by setting $\beta=0$ in Eq. (\ref{eq: maximizing discriminator with F1}), {\it i.e.}, only using the F$_{1-sp}$ scores of ground-truth subsets $S_{DD-I}$ and the fake subsets generated by the initialized $\G_{\boldsymbol{\theta}}$ with $\z$ being the zero vector. The corresponding pre-training parameters are: batch size $=256$, epochs $=20$, learning rate $=1$, $\ell_2$ weight decay $=0.0001$. 
With the initialized $\G_{\boldsymbol{\theta}}$ and the pre-trained $\D_{\boldsymbol{\eta}}$, we fine-tune the D$^2$IA-GAN model using the following parameters: batch size $=256$, epochs $=50$,  the learning rates of $\W$ and $\boldsymbol{\eta}$  are set to $0.0001$ and $0.00005$ respectively, both learning rates are decayed by $0.1$ in every 10 epochs, $\ell_2$ weight decay $=0.0001$, and $\beta=0.5$. 
Besides, if there are a few long paths ({\it i.e.}, many tags in a semantic path) in $SP_{I}$, the number of subsets in $SP_{I}$, {\it i.e.}, $|S_{DD-I}|$, could be very large. In ESP Game and IAPRTC-12, the largest $|SP_{I}|$ is up to 4000, though $|S_{DD-I}|$ for most images are smaller than 30. If $|S_{DD-I}|$ is too large, the training of the discriminator $\D_{\boldsymbol{\eta}}$ (see Eq. (\ref{eq: maximizing discriminator with F1})) will be slow. Thus, we set a upper bound $10$ for $|S_{DD-I}|$ in training, if $|S_{DD-I}| > 10$, then we randomly choose 10 subsets from $S_{DD-I}$ to update $\D_{\boldsymbol{\eta}}$.
The implementation adopts Tensorflow 1.2.0 and Python 2.7.  

\vspace{3pt}
\noindent
{\bf Evaluation metrics.}  
To evaluate the distinctiveness and relevance of the predicted tag subset, three semantic metrics, including semantic precision, recall and F1, are proposed in \cite{my-cvpr-2017-dia}, according to the weighted semantic paths. They are denoted as P$_{sp}$,  R$_{sp}$ and F$_{1-sp}$ respectively. Specifically, given a predicted subset $T$, the corresponding semantic paths $SP_T$ and the ground-truth semantic paths $SP_I$, P$_{sp}$ computes the proportion of the true semantic paths in $SP_T$, and R$_{sp}$ computes the proportion of the true semantic paths in $SP_I$ that are also included in $SP_T$, and F$_{1-sp} = 2(\text{P}_{sp} \cdot \text{R}_{sp}) / (\text{P}_{sp} + \text{R}_{sp})$. The tag weight in each path is also considered when computes the proportion.  
Please refer to \cite{my-cvpr-2017-dia} for the detailed definition.

\vspace{3pt}
\noindent
{\bf Comparisons.} 
We compare with two state-of-the-art image annotation methods, including ML-MG\footnote{Downloaded from {\it https://sites.google.com/site/baoyuanwu2015/home}} \cite{my-iccv-2015} and DIA\footnote{Downloaded from {\it https://github.com/wubaoyuan/DIA}} \cite{my-cvpr-2017-dia}. 
The reason we compare with them is that both of them and the proposed method utilize the semantic hierarchy and the weighted semantic paths, but with different usages. 
We also compare with another state-of-the-art multi-label learning method, called LEML\footnote{Downloaded from {\it http://www.cs.utexas.edu/~rofuyu/}} \cite{LEML-ICML-2014}, which doesn't utilize the semantic hierarchy.
Since both ML-MG and LEML do not consider the semantic distinctiveness among tags, their predicted tag subsets are likely to include semantic redundancies. As reported in \cite{my-cvpr-2017-dia}, the evaluation scores using the semantic metrics ({\it i.e.}, P$_{sp}$, R$_{sp}$ and F$_{1-sp}$) of ML-MG and LEML's predictions are much lower than DIA. Hence it is not relevant to compare with the original results of ML-MG and LEML. Instead, we combine the predictions of ML-MG and LEML with the DPP-sampling that is also used in DIA and our method. Specifically, the square root of posterior probabilities with respect to all candidate tags produced by ML-MG are used as the quality vector (see Section \ref{sec: background}); as there are negative scores in the predictions of LEML, we normalize all predicted scores to $[0,1]$ to obtain the posterior probabilities. Then combining with the similarity matrix $\mathbf{S}$, a DPP distribution is constructed to sampling a distinct tag subset. The obtained results denoted as MLMG-DPP and LEML-DPP respectively.

\subsection{Quantitative Results}
\label{sec: subsec experimental results}

As all compared methods (MLMG-DPP, LEML-DPP and DIA) and the proposed method D$^2$IA-GAN sample DPP models to generate tag subsets, we can generate multiple tag subsets using each method for each image. 
Specifically, MLMG-DPP and DIA generates 10 random tag subsets for each image. The weight of each tag subset is computed by summing the weights of all tags in the subset. 
Then we construct two outputs: the {\it single subset}, which picks the subset with the largest weight from these 10 subsets; and the {\it ensemble subset}, which merges 5 tag subsets with top-5 largest weights among 10 subsets into one unique tag subset. 
The evaluations of the single subset reflect the performance of distinctiveness of the compared methods. 
The evaluations of the ensemble subset measure the performance of both diversity and distinctiveness. Larger distinctiveness of the ensemble subset indicates higher diversity among the consisting subsets of this ensemble subset. Besides, we present two cases by limiting the size of each tag subset to 3 and 5, respectively. 

\renewcommand{\arraystretch}{0.9}
\begin{table}[t] 
\begin{center}
\scalebox{0.74}{
\begin{tabular}{|p{.075\textwidth}|p{.105\textwidth}| p{.045\textwidth} p{.045\textwidth} p{.045\textwidth} | p{.045\textwidth} p{.045\textwidth} p{.045\textwidth} |}
\hline
evaluation & metric$\rightarrow$ &
\multicolumn{3}{|c|}{3 tags} & \multicolumn{3}{|c|}{5 tags}
\\
target & method$\downarrow$ & P$_{sp}$ & R$_{sp}$ & F$_{1-sp}$ & P$_{sp}$ & R$_{sp}$ & F$_{1-sp}$
 \\
 \hline \hline
 & \scalebox{0.8}{LEML-DPP \cite{LEML-ICML-2014}} & 34.64 & 25.21 & 27.76 & 29.24 & 35.05 & 30.29
\\ 
 \scalebox{1}{\multirow{1}{*}{single}} & \scalebox{0.8}{MLMG-DPP \cite{my-iccv-2015}} & 37.18 & 27.71 & 30.05 & 33.85 & 38.91 & 34.30
 \\
 \scalebox{1}{\multirow{1}{*}{subset}} & \scalebox{0.9}{DIA \cite{my-cvpr-2017-dia}} & 41.44 & 31.00 & 33.61 & 34.99 & 40.92 & 35.78
 \\
 & \scalebox{0.9}{D$^2$-GAN} & \textbf{42.96}  & \textbf{32.34} & \textbf{34.93} & \textbf{35.04} & \textbf{41.50} & \textbf{36.06}
 \\
 \hline \hline
 & \scalebox{0.8}{LEML-DPP \cite{LEML-ICML-2014}} & 34.62 & 38.09 & 34.32 & 29.04 & 46.61 & 34.02
\\
 \scalebox{1}{\multirow{1}{*}{ensemble}} & \scalebox{0.8}{MLMG-DPP \cite{my-iccv-2015}} & 30.44 &  34.88 & 30.70 & 28.99 & 43.46 & 33.05 		
 \\
 \scalebox{1}{\multirow{1}{*}{subset}} & \scalebox{0.9}{DIA \cite{my-cvpr-2017-dia}} & 35.73 & 33.53 & 32.39 & \textbf{32.62} & 40.86 & 34.31	
 \\
 & \scalebox{0.9}{D$^2$-GAN}  & \textbf{36.73} & \textbf{42.44} & \textbf{36.71} & 31.28 & \textbf{48.74} & \textbf{35.82}	
 \\
 \hline
\end{tabular}
}
\end{center}
\vspace{-0.05in}
\caption{ Results ($\%$) evaluated by semantic metrics on ESP Game. The higher value indicates the better performance, and the best result in each column is highlighted in bold.}
\label{table: result on espgame}
\vspace{-0.1in}
\end{table}

\renewcommand{\arraystretch}{0.9}
\begin{table}[t] 
\begin{center}
\scalebox{0.74}{
\begin{tabular}{|p{.075\textwidth}|p{.105\textwidth}| p{.045\textwidth} p{.045\textwidth} p{.045\textwidth} | p{.045\textwidth} p{.045\textwidth} p{.045\textwidth} |}
\hline
evaluation & metric$\rightarrow$ &
\multicolumn{3}{|c|}{3 tags} & \multicolumn{3}{|c|}{5 tags}
\\
target & method$\downarrow$ & P$_{sp}$ & R$_{sp}$ & F$_{1-sp}$ & P$_{sp}$ & R$_{sp}$ & F$_{1-sp}$
 \\
 \hline \hline
  & \scalebox{0.8}{LEML-DPP \cite{LEML-ICML-2014}} & 41.42 & 24.39 & 29.00 & 37.06 & 32.86 & 32.98
 \\
 single & \scalebox{0.8}{MLMG-DPP \cite{my-iccv-2015}} & 40.93 & 24.29 & 28.61 & 37.06 & 33.68 & 33.29
 \\
 subset & \scalebox{0.9}{DIA \cite{my-cvpr-2017-dia}} & 42.65 & 25.07 & 29.87 & \textbf{37.83} & 34.62 & 34.11
 \\
 & \scalebox{0.9}{D$^2$-GAN}  & \textbf{43.57} & \textbf{26.22} & \textbf{31.04} & 37.31 & \textbf{35.35} & \textbf{34.41}
 \\
 \hline \hline
 & \scalebox{0.8}{LEML-DPP \cite{LEML-ICML-2014}} & 35.22 & 32.75 & 31.86 & 32.28 & 39.89 & 33.74
\\
 ensemble & \scalebox{0.8}{MLMG-DPP \cite{my-iccv-2015}} & 33.71 & 32.00 & 30.64 & 31.91 & 40.11 & 33.49	
 \\
 subset & \scalebox{0.9}{DIA \cite{my-cvpr-2017-dia}} & \textbf{35.73} & 33.53 & 32.39 & \textbf{32.62} & 40.86 & 34.31 
 \\
 & \scalebox{0.9}{D$^2$-GAN} & 35.49 & \textbf{39.06} & \textbf{34.44} & 32.50 & \textbf{44.98} & \textbf{35.34}
 \\
 \hline
\end{tabular}
}
\end{center}
\vspace{-0.05in}
\caption{ Results ($\%$) evaluated by semantic metrics on IAPRTC-12. The higher value indicates the better performance, and the best result in each column is highlighted in bold.}
\label{table: result on iaprtc12}
\vspace{-0.2in}
\end{table}

The quantitative results on ESP Game are shown in Table \ref{table: result on espgame}. For  single subset evaluations, D$^2$IA-GAN shows the best performance evaluated by all metrics for both 3 and 5 tags, while MLMG-DPP and LEML-DPP perform worst in all cases. The reason is that the learning of ML-MG/LEML and the DPP sampling are independent. For ML-MG, it enforces the ancestor tags to be ranked before its descendant tags, while the distinctiveness is not considered. There is much semantic redundancy in the top-k tags of ML-MG, which is likely to include fewer semantic paths than the ones of DIA and D$^2$IA-GAN. Hence, although DPP sampling can produce a distinct tag subset from the top-k candidate tags, it covers fewer semantic concept (remember that one semantic path represents one semantic concept) than DIA and D$^2$IA-GAN. 
For LEML, it treats each tag equally when training, totally ignoring the semantic distinctiveness. It is not surprising that LEML-DPP also covers fewer semantic concepts than DIA and D$^2$IA-GAN. 
In contrast, both DIA and D$^2$IA-GAN take into account the semantic distinctiveness in learning. 
However, there are several significant differences between their training processes. 
Firstly, the DPP sampling is independent with the model training in DIA, while the generated subset by DPP sampling is used to updated the model parameter in D$^2$IA-GAN. 
Secondly, DIA learns from the ground-truth complete tag list, and the semantic distinctiveness is indirectly embedded into the learning process through the similarity matrix $\mathbf{S}$. In contrast, D$^2$IA-GAN learns from the ground-truth distinct tag subsets. 
Thirdly, the model training of DIA is independent of the evaluation metric F$_{1-sp}$, which plays the important role in the training process of D$^2$IA-GAN. 
These differences are the causes that D$^2$IA-GAN produces more semantically distinct tag subsets than DIA. 
Specifically, in the case of 3 tags, the relative improvements of D$^2$IA-GAN over DIA are $3.67\%, 4.32\%, 3.93\%$ at P$_{sp}$, R$_{sp}$ and F$_{1-sp}$,  respectively; while being $0.14\%, 3.86\%$ and 
 $0.78\%$ in the case of 5 tags. 
In addition, the improvement decreases as the size limit of tag subset increases. The reason is that D$^2$IA-GAN may include more irrelevant tags, as the random noise combined with the image feature not only brings in diversity, but also uncertainty. 
Note that due to the randomness of sampling, the results of single subset by DIA presented here are slightly different with those reported in \cite{my-cvpr-2017-dia}.

In terms of the evaluation of the ensemble subsets, the improvement of D$^2$IA-GAN over three compared methods is more significant. 
This is because all three compared methods sample multiple tag subsets from a fixed DPP distribution, while D$^2$IA-GAN generates multiple tag subsets from different DPP distributions with the random perturbations. 
As such, the diversity among the tag subsets generated by D$^2$IA-GAN is expected to be higher than those corresponding to three compared methods. 
Subsequently, the ensemble subset of D$^2$IA-GAN is likely to cover more relevant semantic paths than those of other methods.
It is supported by the comparison through the evaluation by R$_{sp}$: the relative improvement of D$^2$IA-GAN over DIA is $26.57\%$ in the case of 3 tags, while $19.29\%$ in the case of 5 tags. 
It is encouraging that the P$_{sp}$ scores of D$^2$IA-GAN are also comparable with those of DIA. It demonstrates that training using GAN reduces the likelihood to include irrelevant semantic paths due to the uncertainty of the noise vector $\z$, because GAN encourages the generated tag subsets to be close to the ground-truth diverse and distinct tag subsets. 
Specifically, in the case of 3 tags, the relative improvements of D$^2$IA-GAN over DIA are $2.80\%, 26.57\%, 11.77\%$ for P$_{sp}$, R$_{sp}$ and F$_{1-sp}$, respectively; the corresponding improvements are $-4.11\%, 19.29\%, 4.40\%$ in the case of 5 tags. 

The results on IAPRTC-12 are summarized in Table \ref{table: result on iaprtc12}. 
In the case of single subset with 3 tags, the relative improvements of D$^2$IA-GAN over DIA are $2.16\%, 4.59\%, 3.92\%$ for P$_{sp}$, R$_{sp}$ and F$_{1-sp}$, respectively; 
In the case of single subset with 5 tags, the corresponding improvements are $-1.37\%, 2.11\%, 0.88\%$.
In the case of ensemble subset and 3 tags, the corresponding improvements are $-0.67\%, 16.49\%, 6.33\%$. 
In the case of ensemble subset and 5 tags, the corresponding improvements are $-0.37\%, 10.08\%, 3.0\%$. 
The comparisons on above two benchmark datasets verify that D$^2$IA-GAN produces more semantically diverse and distinct tag subsets than the compared MLMG-DPP and DIA methods.  
Some qualitative results will be presented in the {\bf supplementary material}.

\vspace{-0.1in}
\subsection{Subject Study}
\label{sec: subsec subjet study}
\vspace{-0.05in}

Since the diversity and distinctiveness are subjective concepts,  
we also conduct human subject studies to compare the results of DIA and D$^2$-GAN on these two criterion. 
Specifically, for each test image, we run DIA 10 times to obtain 10 tag subsets, and then the set including 3 subsets with the largest weights are picked as the final output. 
For D$^2$-GAN, we firstly generate 10 random noise vectors $\z$. With each noise vector, we conduct the DPP sampling in $\G_{\boldsymbol{\theta}}$ for 10 times to obtain 10 subsets, out of which we pick the one with the largest weight as the tag subset corresponding to this noise vector. Then from the obtained 10 subsets, we again pick 3 subsets with the largest weights to form the output set of D$^2$-GAN. 
For each test image, we present these two sets of tag subsets with the corresponding image to 5 human evaluators. The only instruction to the subjects is to determine ``which set  describes this image more comprehensively". 
Besides, we notice that if two sets are very similar, or if they both are irrelevant to the image content, human evaluators may pick one randomly. To reduce such randomness, we filter the test images using the following criterion: firstly we combine the subsets in each set to an ensemble subset; if the F$_{1-sp}$ scores of both ensemble subsets are larger than 0.2, and the gap between this two scores is larger than $0.15$, then this image is used in subject studies. 
Finally, the numbers of test images used in subject studies are: ESP Game, $375$ in the case of 3 tags, and $324$ in the case of of 5 tags; IAPRTC-12, $342$ in the case of 3 tags, and $306$ in the case of of 5 tags. 
We also present the comparison results using the F$_{1-sp}$ to evaluate the compared two ensemble subsets. The consistency between the F$_{1-sp}$ evaluation and the human evaluation is also computed. 
The subject study results on ESP Game are summarized in Table \ref{table: subject study on espgame}. With human evaluation, D$^2$IA-GAN is judged better at $\frac{240}{375}=64\%$ of all evaluated images over DIA in the case of 3 tags, and $\frac{204}{324}=62.96\%$ in the case of 5 tags. 
With F$_{1-sp}$ evaluation, D$^2$IA-GAN outperforms DIA at $\frac{250}{375}=66.67\%$ in the case of 3 tags, and $\frac{212}{324}=65.43\%$ in the case of5 tags. 
Both evaluation results suggest the improvement of D$^2$IA-GAN over DIA. 
Besides, the results of these two evaluations are consistent ({\it i.e.}, their decisions of which set is better are same) at $\frac{239}{375}=63.73\%$ of all evaluated images of the case of 3 tags, while $\frac{222}{324}=68.52\%$ of the case of 5 tags. 
It demonstrates that the evaluation using F$_{1-sp}$ is relatively reliable. 
%
The same trend is also observed for the results obtained on the IAPRTC-12 dataset (Table \ref{table: subject study on iaprtc12}). 

Moreover, in the {\bf supplementary material}, we will present a detailed analysis about human annotations conducted on partial images of IAPRTC-12. 
It not only shows that D$^2$IA-GAN produces more human-like tags than DIA, but also discusses the difference between D$^2$IA-GAN and human annotators, and how to shrink that difference.

\renewcommand{\arraystretch}{1}
\begin{table}[t] 
\begin{center}
\scalebox{0.87}{
\begin{tabular}{|p{.106\textwidth}|p{.03\textwidth} p{.05\textwidth} p{.053\textwidth} | p{.03\textwidth} p{.05\textwidth} p{.053\textwidth} |}
\hline
\scalebox{0.7}{$\#$} tags $\rightarrow$ & \multicolumn{3}{c|}{3 tags} & \multicolumn{3}{c|}{5 tags} 
\\
\hline
\multirow{2}{*}{metric $\downarrow$} 
& \scalebox{0.8}{DIA}  & \scalebox{0.7}{D$^2$IA-GAN}  & \hspace{0.5em} \multirow{2}{*}{\scalebox{1}{total}} 
& \scalebox{0.8}{DIA}  & \scalebox{0.7}{D$^2$IA-GAN}  & \hspace{0.5em} \multirow{2}{*}{\scalebox{1}{total}}
 \\
  & \scalebox{0.75}{wins} & \hspace{0.55em} \scalebox{0.75}{wins} &  & \scalebox{0.75}{wins} & \hspace{0.55em} \scalebox{0.75}{wins} &
 \\
 \hline
 \scalebox{0.8}{human evaluation} 
 & 135 & \hspace{0.5em} 240 & \hspace{0.55em} 375 
 & 120 & \hspace{0.5em} 204 & \hspace{0.55em} 324
\\
  F$_{1-sp}$ & 125 & \hspace{0.5em} 250 & \hspace{0.55em} 375  
  & 112 & \hspace{0.5em} 212 & \hspace{0.55em} 324  
 \\
 consistency & 62 & \hspace{0.5em} 177 & $63.73\%$
 & 65 & \hspace{0.5em} 157 & $68.52\%$
 \\
 \hline
\end{tabular}
}
\end{center}
\vspace{-0.06in}
\caption{ \small{\small{Subject study results on ESP Game. Note that the entry `62' corresponding to the row `consistency' and the column `DIA wins' indicates that both human evaluation and F$_{1-sp}$ evaluation decide that the predicted tags of DIA are better than those of D$^2$IA-GAN at 62 images. Similarly, human evaluation and F$_{1-sp}$ evaluation have the same decision that the results of D$^2$IA-GAN are better than those of DIA at 177 images. Hence, two evaluations have the same decision ({\it i.e.}, consistent) on $62+177=239$ images, and the consistency rate among all evaluated images are $239/372=63.73\%$.}}
}
\label{table: subject study on espgame}
\vspace{-0.08in}
\end{table}

\vspace{-0.05in}
\renewcommand{\arraystretch}{1}
\begin{table}[t] 
\begin{center}
\scalebox{0.87}{
\begin{tabular}{|p{.106\textwidth}|p{.03\textwidth} p{.05\textwidth} p{.053\textwidth} | p{.03\textwidth} p{.05\textwidth} p{.053\textwidth} |}
\hline
\scalebox{0.7}{$\#$} tags $\rightarrow$ & \multicolumn{3}{c|}{3 tags} & \multicolumn{3}{c|}{5 tags} 
\\
\hline
\multirow{2}{*}{metric $\downarrow$} 
& \scalebox{0.8}{DIA}  & \scalebox{0.7}{D$^2$IA-GAN}  & \hspace{0.5em} \multirow{2}{*}{\scalebox{1}{total}} 
& \scalebox{0.8}{DIA}  & \scalebox{0.7}{D$^2$IA-GAN}  & \hspace{0.5em} \multirow{2}{*}{\scalebox{1}{total}}
 \\
  & \scalebox{0.75}{wins} & \hspace{0.55em} \scalebox{0.75}{wins} &  & \scalebox{0.75}{wins} & \hspace{0.55em} \scalebox{0.75}{wins} &
 \\
 \hline
 \scalebox{0.8}{human evaluation} 
 & 129 & \hspace{0.5em} 213 & \hspace{0.52em} 342 
 & 123 & \hspace{0.5em} 183 & \hspace{0.52em} 306
\\
  F$_{1-sp}$ & 141 & \hspace{0.5em} 201 & \hspace{0.52em} 342  
  & 123 & \hspace{0.5em} 183 & \hspace{0.52em} 306  
 \\
 consistency & 82 & \hspace{0.5em} 154 & $69.01\%$
 & 58 & \hspace{0.5em} 118 & $57.52\%$
 \\
 \hline
\end{tabular}
}
\end{center}
\vspace{-0.07in}
\caption{ Subject study results on IAPRTC-12.
}
\label{table: subject study on iaprtc12}
\vspace{-0.2in}
\end{table}
\vspace{-0.05in}

\section{Conclusion}
\label{sec: conclusion}
\vspace{-0.05in}

In this work, we have proposed a new image annotation method, called {\it diverse and distinct image annotation} (D$^2$IA), to simulate the diversity and distinctiveness of the tags generated by human annotators. 
D$^2$IA is formulated as a sequential generative model, in which the image feature is firstly incorporated into a determinantal point process (DPP) model that also encodes the weighted semantic paths, from which a sequence of distinct tags are generated by sampling. 
The diversity among the generated multiple tag subsets is ensured by sampling the DPP model with random noise perturbations to the image feature.
In addition, we adopt the generative adversarial network (GAN) model to train the generative model D$^2$IA, and employ the policy gradient algorithm to handle the training difficulty due to the discrete DPP sampling in D$^2$IA. 
Experimental results and human subject studies on benchmark datasets demonstrate that the diverse and distinct tag subsets generated by the proposed method can provide more comprehensive descriptions of the image contents than those generated by the state-of-the-art methods.

\vspace{0.05in}
\noindent
\textbf{Acknowledgements}:
This work is supported by Tencent AI Lab. The participation of Bernard Ghanem is supported by the King Abdullah University of Science and Technology (KAUST) Office of Sponsored Research.  
The participation of Siwei Lyu is partially supported by National Science Foundation National Robotics Initiative (NRI) Grant (IIS-1537257) and National Science Foundation of China Project Number 61771341.

{\small
\bibliographystyle{ieee}
\bibliography{bywu_bib}
}

\end{document}